\documentclass{article}
\usepackage{spconf,amsmath,graphicx,amssymb,booktabs,multirow,array,makecell,xspace,bm}

\title{STAN: SPATIO-TEMPORAL ADVERSARIAL NETWORKS\\FOR ABNORMAL EVENT DETECTION}
\name{Sangmin Lee, Hak Gu Kim, and Yong Man Ro*\thanks{* Corresponding author (ymro@ee.kaist.ac.kr)}}
\address{Image and Video Systems Lab., School of Electrical Engineering, KAIST, South Korea}
\begin{document}
%
\maketitle
\begin{abstract}
In this paper, we propose a novel abnormal event detection method with spatio-temporal adversarial networks (STAN). We devise a spatio-temporal generator which synthesizes an inter-frame by considering spatio-temporal characteristics with bidirectional ConvLSTM. A proposed spatio-temporal discriminator determines whether an input sequence is real-normal or not with 3D convolutional layers. These two networks are trained in an adversarial way to effectively encode spatio-temporal features of normal patterns. After the learning, the generator and the discriminator can be independently used as detectors, and deviations from the learned normal patterns are detected as abnormalities. Experimental results show that the proposed method achieved competitive performance compared to the state-of-the-art methods. Further, for the interpretation, we visualize the location of abnormal events detected by the proposed networks using a generator loss and discriminator gradients.
\end{abstract}
\section{Introduction}
\label{sec:1}

With a rapid development of storage and video acquisition devices, the video surveillance system is widely used for privacy protection, process monitoring in factories, criminal tracking, etc. Most of the surveillance scenes captured for a long time are meaningless and normal. It is time-consuming and labor-intensive for people to watch long hours of meaningless scenes. To address the problem, automatic abnormal event (\emph{i}.\emph{e}., meaningful moment) detection in videos has increasingly attracted attention in image processing and computer vision fields.

There were many existing methods for automatically detecting abnormal events in videos. In \cite{1,3}, trajectory based abnormal event detection methods which utilize the dynamic information of objects were proposed. Hu \emph{et al}.\ \cite{1} proposed method for learning statistical motion patterns with multi-object tracking. Zhou \emph{et al}.\ \cite{3} proposed statistical model to detect abnormal behaviors using Kanade-Lucas-Tomasi Feature Tracker (KLT). However, trajectory based methods are not robust to occlusions and crowded scenes since most of the object tracking algorithms are vulnerable to such environments. In \cite{4,5}, handcraft feature based abnormal event detection methods were proposed. Kaltsa \emph{et al}.\ \cite{4} utilized a descriptor created from Histograms of Oriented Gradients (HOG) and Histograms of Oriented Swarms (HOS). In \cite{5}, distributions of spatio-temporal oriented energy were used to model behavior. Although the handcraft feature based methods are more robust to occlusions and complex scenes than trajectory based methods, they require prior knowledge to design proper features for various events.

In recent years, deep learning has attracted attention in many challenging tasks with better performance compared to the traditional methods \cite{6,7,8}. In this flow, several methods using deep learning have been proposed for abnormal event detection. Hasan \emph{et al}.\ \cite{9} proposed the convolutional autoencoder based method for learning temporal regularity. In \cite{10}, the ConvLSTM \cite{23} was employed to consider the spatio-temporal characteristics of event patterns. Recently, generative adversarial networks (GAN) based method \cite{11} was proposed. In \cite{11}, one GAN is trained to generate an optical flow map from a frame while the other GAN is trained to generate a frame from an optical flow map. Although the method of \cite{11} achieved good performance by using GAN, its performance could be sensitive to the quality of the estimated optical flow map. It could not be robust to scenes with occlusions in which it is difficult to well estimate the optical flow map.

In this paper, we propose the spatio-temporal adversarial networks (STAN) which learn the spatio-temporal features of normal patterns. Learning the characteristics of abnormal events is very difficult because the definition of such events is not spatially or temporally bounded. Thus, it is reasonable to learn normal patterns and consider deviations from the learned patterns as abnormalities. With the proposed STAN, we directly generate spatio-temporal scene without using the optical flow map. The STAN contains two networks which are the spatio-temporal generator and the spatio-temporal discriminator. The proposed generator consists of three network modules which are a spatial encoder for encoding spatial features of frames, a bidirectional ConvLSTM for encoding temporal feature of the scene, and a spatial decoder for generating an inter-frame. By using the bidirectional configuration, the spatio-temporal features of normal patterns can be encoded in forward and backward directions. The proposed discriminator is a 3D convolutional neural network for determining whether an input sequence is real or not.

\begin{figure}[t]
	\begin{minipage}[b]{1.0\linewidth}
		\centering
		\centerline{\includegraphics[width=8.5cm]{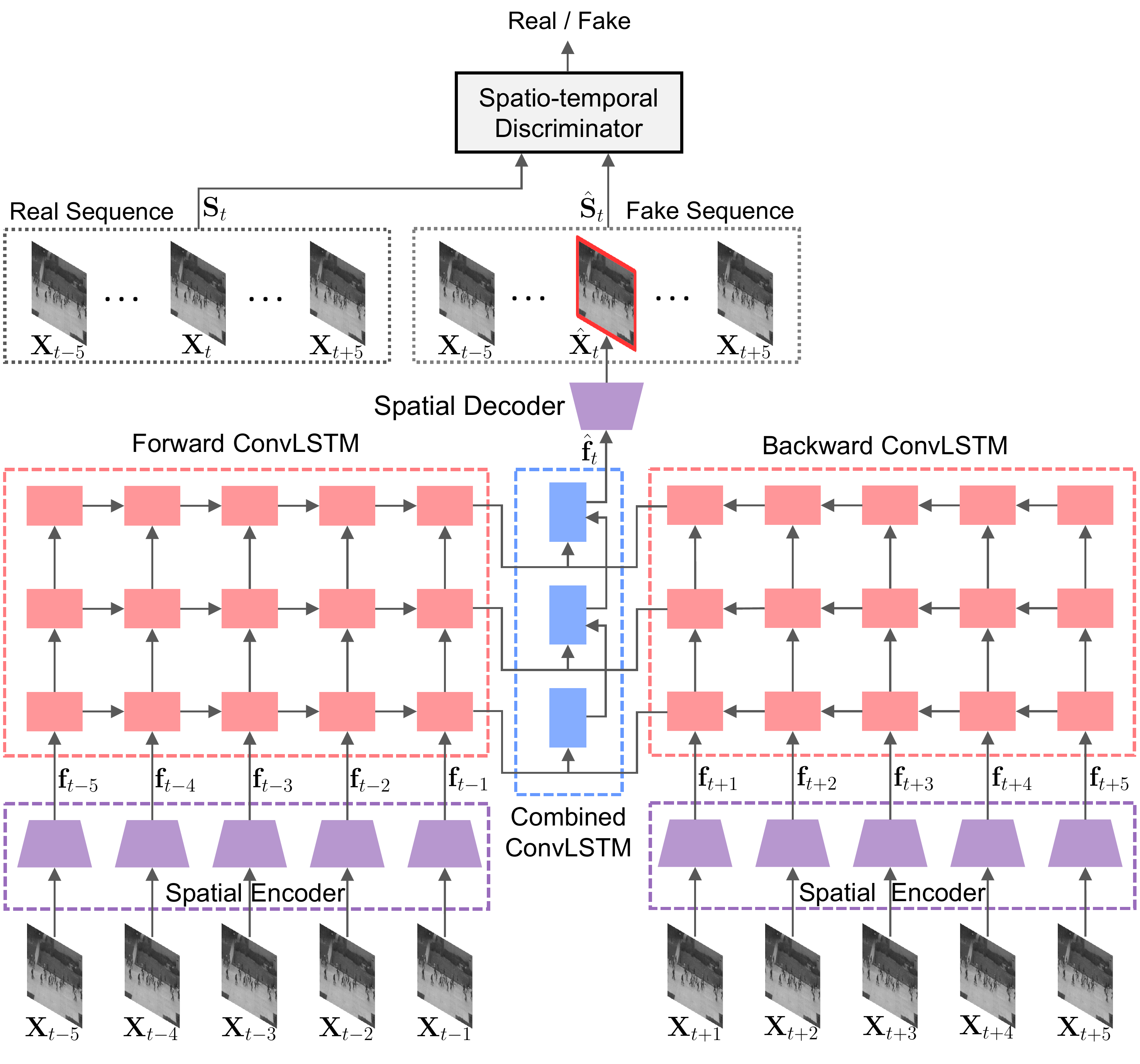}}
	\end{minipage}
	\vspace{-0.8cm}
	\caption{Spatio-temporal adversarial networks at training time.}
	\label{fig:1}
	\vspace{-0.3cm}
\end{figure}

By the adversarial learning, the generator tries to generate the inter-frame in order to deceive the discriminator while the discriminator tries to discriminate between a real sequence and a fake sequence. Note that the fake sequence includes the generated inter-frame. At testing time, abnormal events are detected by using a pixel-wise loss for the generator and an adversarial loss for the discriminator.

Experimental results show that the proposed method performs competitively compared to the state-of-the-art methods, especially on complex scenes with large objects and frequent occlusions. In addition, we visually interpret the proposed networks by locating detected abnormal events through both the generator and the discriminator.

The remainder of this paper is organized as follows. Section 2 describes the proposed method in detail. In Section 3, experimental results are shown for qualitative and quantitative evaluation. Finally, conclusions are drawn in Section 4.

\section{Proposed method}
\label{sec:2}

Fig.~\ref{fig:1} shows the overall procedure of the proposed STAN at training time. The proposed spatio-temporal generator generates the inter-frame by considering neighboring frames. Then, consecutive frames including the generated inter-frame are considered as a fake sequence while a real sequence contains only real frames. Through the adversarial learning, the spatio-temporal discriminator is trained to determine whether the input sequence is real or fake. The generator is trained to generate the inter-frame that can fool the discriminator. Details are described in following sub-sections.

\subsection{Spatio-temporal generator}
\label{ssec:2.1}

The role of the proposed generator is to generate a clear frame for the normal scene and a distorted frame for the abnormal scene. In video data, each frame has a high correlation with neighboring frames. We devise the bidirectional ConvLSTM to generate the inter-frame by considering the spatio-temporal characteristics of the neighboring frames. With the bidirectional configuration, the spatio-temporal features of normal patterns are encoded effectively. As a result, scenes with spatio-temporal abnormal patterns can be sensitively detected.

The proposed generator consists of the spatial encoder, the bidirectional ConvLSTM, and the spatial decoder. By the spatial encoder, latent spatial features are encoded for representing visual information of each frame. The spatial features of the previous 5 frames and the later 5 frames are fed to the forward ConvLSTM and the backward ConvLSTM respectively, in order to encode the spatio-temporal features. Then, the hidden states and the cell states resulting from each ConvLSTM module are concatenated in pairs to combine forward and backward features. As shown in Fig.~\ref{fig:1}, the combined ConvLSTM outputs the latent feature of the inter-frame. Finally, the inter-frame is generated by the spatial decoder.

To generate the desirable inter-frame, a realism loss and a pixel-wise loss are designed for the generator. By minimizing the realism loss, the generator can force the discriminator to fail to classify the fake sequence as fake. Let $G_{\boldsymbol{\theta}}$ and $D_{\boldsymbol{\phi}}$ denote the generator function with parameter ${\boldsymbol{\theta}}$ and the discriminator function with parameter ${\boldsymbol{\phi}}$, respectively. Let $\mathrm{\mathbf{X}}_t$ $\in$ $\mathbb{R}^{224\times224\times1}$  and $\mathrm{\hat{\mathbf{X}}}_t$ $\in$ $\mathbb{R}^{224\times224\times1}$ denote the $t$-th real frame and the $t$-th generated frame by $G_{\boldsymbol{\theta}}$, respectively. The realism loss can be written as 
\begin{align}
\ell_{\mathrm{real}}({\boldsymbol{\theta}};t)=-\log(D_{\boldsymbol{\phi}}(\mathrm{\hat{\mathbf{S}}}_t)),
\label{eq:1}
\end{align}
\noindent where ${{\hat{\mathbf{S}}}}_t$ $\in$ $\mathbb{R}^{11\times224\times224\times1}$ is the fake sequence including $t$-th generated frame ${\hat{\mathbf{X}}}_t$.

In addition, by minimizing the pixel-wise loss between the $t$-th real frame $\mathrm{\mathbf{X}}_t$ and the $t$-th generated frame $\mathrm{\hat{\mathbf{X}}}_t$, the generated frame resembles the real frame in pixel-level. The pixel-wise loss can be written as
\begin{align}
\ell_{\mathrm{pixel}}({\boldsymbol{\theta}};t)=\|G_{\boldsymbol{\theta}}(\mathrm{{\mathbf{X}}}_{t-5}, ...,\mathrm{{\mathbf{X}}}_{t-1}, \mathrm{{\mathbf{X}}}_{t+1}, ..., \mathrm{{\mathbf{X}}}_{t+5})-\mathrm{\mathbf{X}}_t\|_2.
\label{eq:2}
\end{align}

Finally, the total loss for the generator is defined as a combination of the two losses, which can be written as
\begin{align}
\mathcal{L}_G({\boldsymbol{\theta}})=\sum_{t \in batch}\ell_{\mathrm{real}}({\boldsymbol{\theta}};t)+\lambda\ell_{\mathrm{pixel}}({\boldsymbol{\theta}};t),
\label{eq:3}
\end{align}
\noindent where $\lambda$ is a hyper-parameter to balance the realism loss and the pixel-wise loss.
\newcolumntype{P}[1]{>{\centering\arraybackslash}p{#1}}
\newcolumntype{M}[1]{>{\centering\arraybackslash}m{#1}}
\begin{table}[t!]
	\vspace{-0.3cm}
	\caption{Architectures of the proposed networks.} 
	\begin{center}
		\small
		\resizebox{0.9999\linewidth}{!}{
			\begin{tabular}{M{1.5cm}M{0.8cm}M{1.8cm}M{0.1cm}M{1.4cm}M{1.1cm}M{2.1cm}}
				\toprule
				
				\multicolumn{3}{c}{\textbf{Generator}} && \multicolumn{3}{c}{\textbf{Discriminator}} \\
				\cmidrule(l){1-3}
				\cmidrule(l){7-7}
				\cmidrule(r){5-7}
				\cmidrule(r){1-1}
				\multirow{2}{*}{\textbf{Layer}} & \textbf{Filter/} & \textbf{Output Size} && \multirow{2}{*}{\textbf{Layer}} & \textbf{Filter/} & \textbf{Output Size} \\
				& \textbf{Stride} & \textbf{($\bm{h}$$\bm{\times}$$\bm{w}$$\bm{\times}$$\bm{c}$)} &&  & \textbf{Stride} & \textbf{($\bm{l}$$\bm{\times}$$\bm{h}$$\bm{\times}$$\bm{w}$$\bm{\times}$$\bm{c}$)} \\ \toprule
				
				\multirow{2}{*}{Conv1} & $5$$\times$$5$/ & \multirow{2}{*}{112$\times$112$\times$16} &&  &  &  \\
				& (2, 2) &  && \multirow{2}{*}{3D Conv1} & \multirow{1}{*}{5$\times$5$\times$5/} & \multirow{2}{*}{7$\times$112$\times$112$\times$32} \\
				\multirow{2}{*}{Conv2} & $5$$\times$$5$/ & \multirow{2}{*}{56$\times$56$\times$32} &&   & \multirow{1}{*}{(1, 2, 2)} &   \\
				& (2, 2) &  &&  &  &  \\ \midrule
				
				\multirow{2}{*}{Conv3} & $3$$\times$$3$/ & \multirow{2}{*}{28$\times$28$\times$64} &&  &  &  \\
				& (2, 2) &  && \multirow{2}{*}{3D Conv1} & \multirow{1}{*}{3$\times$5$\times$5/} & \multirow{2}{*}{5$\times$56$\times$56$\times$64} \\
				\multirow{2}{*}{Conv4} & $3$$\times$$3$/ & \multirow{2}{*}{28$\times$28$\times$128} &&   & \multirow{1}{*}{(1, 2, 2)} &   \\
				& (1, 1) &  &&  &  &  \\ \midrule
				
				& &  &&  &  &  \\
				\multirow{1}{*}{Forward} & \multirow{1}{*}{3$\times$3/} & \multirow{2}{*}{28$\times$28$\times$64}  && \multirow{2}{*}{3D Conv3} & \multirow{1}{*}{3$\times$3$\times$3/} & \multirow{2}{*}{3$\times$28$\times$28$\times$128} \\
				\multirow{1}{*}{ConvLSTM} & (1, 1) &  &&  & \multirow{1}{*}{(1, 2, 2)} &   \\
				&  &  &&  &  &  \\ \midrule
				
				& &  &&  &  &  \\
				\multirow{1}{*}{Backward} & \multirow{1}{*}{3$\times$3/} & \multirow{2}{*}{28$\times$28$\times$64}  && \multirow{2}{*}{3D Conv4} & \multirow{1}{*}{3$\times$3$\times$3/} & \multirow{2}{*}{1$\times$14$\times$14$\times$256} \\
				\multirow{1}{*}{ConvLSTM} & (1, 1) &  &&  & \multirow{1}{*}{(1, 2, 2)} &   \\
				&  &  &&  &  &  \\ \midrule
				
				& &  &&  &  &  \\
				\multirow{1}{*}{Combined} & \multirow{1}{*}{3$\times$3/} & \multirow{2}{*}{28$\times$28$\times$128}  && \multirow{2}{*}{3D Conv5} & \multirow{1}{*}{1$\times$3$\times$3/} & \multirow{2}{*}{1$\times$7$\times$7$\times$512} \\
				\multirow{1}{*}{ConvLSTM} & (1, 1) &  &&  & \multirow{1}{*}{(1, 2, 2)} &   \\
				&  &  &&  &  &  \\ \midrule
				
				\multirow{2}{*}{DeConv1} & $3$$\times$$3$/ & \multirow{2}{*}{28$\times$28$\times$64} &&  &  &  \\
				& (1, 1) &  && \multirow{2}{*}{3D Conv5} & \multirow{1}{*}{1$\times$3$\times$3/} & \multirow{2}{*}{1$\times$7$\times$7$\times$1} \\
				\multirow{2}{*}{DeConv2} & $3$$\times$$3$/ & \multirow{2}{*}{56$\times$56$\times$32} &&   & \multirow{1}{*}{(1, 1, 1)} &   \\
				& (2, 2) &  &&  &  &  \\ \midrule
				
				\multirow{2}{*}{DeConv3} & $5$$\times$$5$/ & \multirow{2}{*}{112$\times$112$\times$16} &&  &  &  \\
				& (2, 2) &  && &  &  \\
				\multirow{2}{*}{DeConv4} & $5$$\times$$5$/ & \multirow{2}{*}{224$\times$224$\times$1} &&   &  &   \\
				& (2, 2) &  &&  &  &  \\ \bottomrule
		\end{tabular}}
	\end{center}
	\label{table:1}
	\vspace{-0.5cm}
\end{table}

\subsection{Spatio-temporal discriminator}
\label{ssec:2.2}

The proposed spatio-temporal discriminator consists of 3D convolutional layers. The 3D CNN can reliably determine whether the input sequence is real or fake by considering both spatial and temporal characteristics of the scene \cite{12}. The discriminator outputs $7\times7$ probability map for employing the local adversarial loss \cite{13}. 

To distinguish the real sequence and the fake sequence, an adversarial loss is designed for the discriminator. By minimizing the loss, the discriminator can discriminate between the real sequence and the fake sequence. The adversarial loss for the discriminator can be written as
\begin{align}
\mathcal{L}_D({\boldsymbol{\phi}})=\sum_{t \in batch}-\log(1-D_{\boldsymbol{\phi}}(\mathrm{\hat{\mathbf{S}}}_t))-\log(D_{\boldsymbol{\phi}}(\mathrm{\mathbf{S}}_t)),
\label{eq:4}
\end{align}
\noindent where $\mathrm{\mathbf{S}}_t$ $\in$ $\mathbb{R}^{11\times224\times224\times1}$ is the real sequence. The first term, $-\log(1-D_{\boldsymbol{\phi}}(\mathrm{\hat{\mathbf{S}}}_t))$, allows the discriminator to classify the fake sequence as fake. The second term, $-\log(D_{\boldsymbol{\phi}}(\mathrm{\mathbf{S}}_t))$, allows the discriminator to classify the real sequence as real.

The proposed discriminator plays two roles. First, during the adversarial learning, the discriminator helps the generator learn the spatio-temporal features of normal patterns. Second, after the adversarial learning, the discriminator alone can detect abnormal events. The discriminator is trained to determine only real sequence containing normal events as real. Thus, the discriminator considers the real sequence containing abnormal events as not real. Consequently, the second term in~\eqref{eq:4} has a higher value for video sequences including abnormal events and can be used to detect abnormalities. Table~\ref{table:1} shows architecture details of the networks.

\subsection{Abnormality score}
\label{ssec:2.3}

To quantify detected abnormalities, we devise a novel abnormality score by using the losses of the generator and the discriminator. The generator and the discriminator complement detection results of each other. Thereby, robust abnormality detection can be achieved. The proposed abnormality loss can be defined as
\begin{align}
\begin{split}
\ell_s(t)=&\|G_{\boldsymbol{\theta}}(\mathrm{{\mathbf{X}}}_{t-5}, ...,\mathrm{{\mathbf{X}}}_{t-1}, \mathrm{{\mathbf{X}}}_{t+1}, ..., \mathrm{{\mathbf{X}}}_{t+5})-\mathrm{\mathbf{X}}_t\|_2 \\
&-\lambda_s\log(D_{\boldsymbol{\phi}}(\mathrm{\mathbf{S}}_t)),
\end{split}
\label{eq:5}
\end{align}
\noindent where $\lambda_s$ is a hyper-parameter to balance the generator detection and the discriminator detection.

Then, by normalizing $\ell_s(t)$, abnormality score $s(t)$ at $t$-th frame can be written as
\begin{align}
s(t)=\frac{\ell_s(t)-\min_{t}\ell_s(t)}{\max_{t}\ell_s(t)-\min_{t}\ell_s(t)}.
\label{eq:6}
\end{align}
\noindent Note that video sequences containing abnormal events have higher abnormality scores. 

\section{Experiments and results}
\label{sec:3}

\subsection{Dataset}
\label{ssec:3.1}

For experiments, we used three datasets: UCSD Ped1 \cite{14}, UCSD Ped2 \cite{14} and Avenue \cite{15} datasets. The training sets of these datasets contain only normal events while the testing sets contain both normal and abnormal events. The UCSD Ped1 dataset consists of 34 clips for training and 36 clips for testing. The UCSD Ped2 dataset consists of 16 clips for training and 12 clips for testing. The Avenue dataset consists of 16 clips for training and 21 clips for testing. The UCSD Ped1 and Ped2 datasets contain the motion of small objects in a broad area. On the other hand, the Avenue dataset contains complex motion of large objects and frequent occlusions.

\subsection{Implementation details}
\label{ssec:3.2}

We used the Adam \cite{24} to optimize the networks with a learning rate of 0.0002 and a batch size of 3. At training time, first, only the generator was trained to minimize the pixel-wise loss. Then, the generator and the discriminator were trained in the adversarial way to minimize $\mathcal{L}_G({\boldsymbol{\theta}})$ and $\mathcal{L}_D({\boldsymbol{\phi}})$ alternately. We set $\lambda$ as 1 in~\eqref{eq:3} and $\lambda_s$ as maximum value ratio divided by 10 of the two terms in~\eqref{eq:5}. At the end of the generator and the discriminator, the tanh and the sigmoid were used as activation functions respectively. For all the other parts, the exponential linear unit (ELU) was used.

\newcolumntype{P}[1]{>{\centering\arraybackslash}p{#1}}
\newcolumntype{M}[1]{>{\centering\arraybackslash}m{#1}}
\begin{table}[t!]
	\vspace{-0.3cm}
	\caption{Frame-level performance of the proposed method.} 
	\begin{center}
		\small
		\resizebox{0.9999\linewidth}{!}{
			\begin{tabular}{M{4.3cm}M{1.5cm}M{1.5cm}M{1.5cm}}
				\toprule

				\multirow{3}{*}{\textbf{Method}} & \multicolumn{3}{c}{\multirow{1}{*}{\textbf{\ AUC (\%)}}} \\ \cmidrule(lr){2-4}
				& \multirow{1}{*}{\textbf{UCSD}} & \multirow{1}{*}{\textbf{UCSD}} & \multirow{2}{*}{\textbf{Avenue}} \\
				& \multirow{1}{*}{\textbf{Ped1}} & \multirow{1}{*}{\textbf{Ped2}} &  \\ \toprule
				
				MPPCA \cite{16} & 59.0 & 69.3 & N / A \\
				Social force (SF) \cite{17} & 67.5 & 55.6 & N / A \\
				MPPCA + SF \cite{14} & 68.8 & 61.3 & N / A \\
				Detection at 150fps \cite{15}
& 91.8 & N / A  & 80.9 \\ \midrule
				
				AMDN \cite{18} & 92.1 & 90.8 & N / A \\
				Conv-AE \cite{9} & 81.0 & 90.0 & 70.2 \\ 
				ConvLSTM-AE \cite{19} & 75.5 & 88.1 & 77.0 \\ 
				Unmasking \cite{20} & 68.4 & 82.2 & 80.6\\ 
				Optical flow-GAN \cite{11} & \textbf{97.4} & 93.5 & N / A \\ 
				Stacked RNN \cite{21} & N / A  & 92.2 & 81.7 \\ \midrule
				
				\textbf{Proposed method ($\boldsymbol{G_{\boldsymbol{\theta}}}$)} & 81.6 & 95.9 & 86.6\\ 
				\textbf{Proposed method ($\boldsymbol{D_{\boldsymbol{\phi}}}$)} & 71.6 & 86.2 & 81.8 \\ 
				\textbf{Proposed method ($\boldsymbol{G_{\boldsymbol{\theta}}}$ and $\boldsymbol{D_{\boldsymbol{\phi}}}$)} & 82.1  & \textbf{96.5} & \textbf{87.2} \\ \bottomrule
						
		\end{tabular}}
	\end{center}
	\label{table:2}
	\vspace{-0.55cm}
\end{table}

\newcolumntype{P}[1]{>{\centering\arraybackslash}p{#1}}
\newcolumntype{M}[1]{>{\centering\arraybackslash}m{#1}}
\begin{table}[t!]
	\caption{Event-level performance of the proposed method.} 
	\begin{center}
		\small
		\resizebox{0.9999\linewidth}{!}{
			\begin{tabular}{M{4.3cm}M{0.99cm}M{0.99cm}M{0.99cm}M{0.99cm}}
				\toprule

				\multirow{4}{*}{\textbf{Method}} & \multicolumn{2}{c}{\multirow{1}{*}{\textbf{Correct Detection /}}} & \multicolumn{2}{c}{\multirow{2}{*}{\textbf{Precision (\%)}}} \\ 
				& \multicolumn{2}{c}{\multirow{1}{*}{\textbf{False Alarm}}} && \\ \cmidrule(lr){2-3} \cmidrule(lr){4-5}
				& \multicolumn{1}{c}{\multirow{1}{*}{\textbf{UCSD}}} & \multicolumn{1}{c}{\multirow{1}{*}{\textbf{UCSD}}} & \multicolumn{1}{c}{\multirow{1}{*}{\textbf{UCSD}}} & \multicolumn{1}{c}{\multirow{1}{*}{\textbf{UCSD}}} \\
				& \multicolumn{1}{c}{\multirow{1}{*}{\textbf{Ped1}}} & \multicolumn{1}{c}{\multirow{1}{*}{\textbf{Ped2}}} & \multicolumn{1}{c}{\multirow{1}{*}{\textbf{Ped1}}} & \multicolumn{1}{c}{\multirow{1}{*}{\textbf{Ped2}}} \\ \toprule
				
				IT-AE \cite{9} & 36 / 11 & 12 / 3 & 76.6 & 80.0 \\
				Conv-AE \cite{9} & 38 / 6 & 12 / 1 & 86.3 & 92.3 \\
				\textbf{Proposed method ($\boldsymbol{G_{\boldsymbol{\theta}}}$ and $\boldsymbol{D_{\boldsymbol{\phi}}}$)} & \textbf{37 / 3} & \textbf{12 / 0} & \textbf{92.5} & \textbf{100.0} \\ \bottomrule

		\end{tabular}}
	\end{center}
	\label{table:3}
	\vspace{-0.5cm}
\end{table}

\subsection{Performance evaluation}
\label{ssec:3.3}

To evaluate the performance of the proposed method, two evaluation metrics were employed. First, we employed the frame-level evaluation \cite{25} based on the area under curves (AUC). Second, we employed event-level evaluation \cite{9} based on the number of detected events. 

\begin{figure}[t]
	\begin{minipage}[b]{1.0\linewidth}
		\centering
		\centerline{\includegraphics[width=8.5cm]{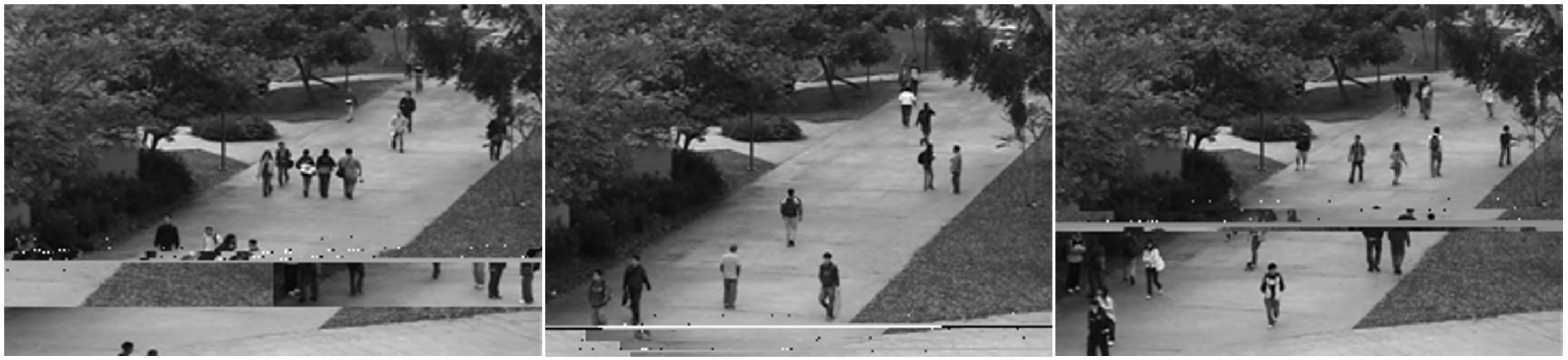}}
	\end{minipage}
	\vspace{-0.8cm}
	\caption{Examples of the corrupted frames in UCSD Ped1.}
	\label{fig:2}
\end{figure}

Table~\ref{table:2} shows frame-level performance compared with handcraft and sparsity based methods \cite{14,15,16,17} and deep learning based methods \cite{9,11,18,19,20,21}. The AUC values of the other methods were taken from each paper and \cite{18}. Interestingly, the proposed method even with ${G_{\boldsymbol{\theta}}}$ or ${D_{\boldsymbol{\phi}}}$ showed respectably high AUC values on the Avenue. The proposed method with both ${G_{\boldsymbol{\theta}}}$ and ${D_{\boldsymbol{\phi}}}$ outperformed the other methods including state-of-the-art methods. On the other hand, the performance on the UCSD Ped1 showed a bit lower than that on the other datasets. We observed that the UCSD Ped1 included several corrupted frames in testing clips as shown in Fig.~\ref{fig:2}. By the proposed method, these corrupted frames were detected as abnormal events even though such frames were denoted as normal events in ground truth labels.

Table~\ref{table:3} shows event-level performance comparisons with correct detection, false alarm, and precision. The proposed method achieved higher precision results compared to the other methods in event-level evaluation.

\begin{figure}[t]
	\begin{minipage}[b]{1.0\linewidth}
		\centering
		\centerline{\includegraphics[width=8.5cm]{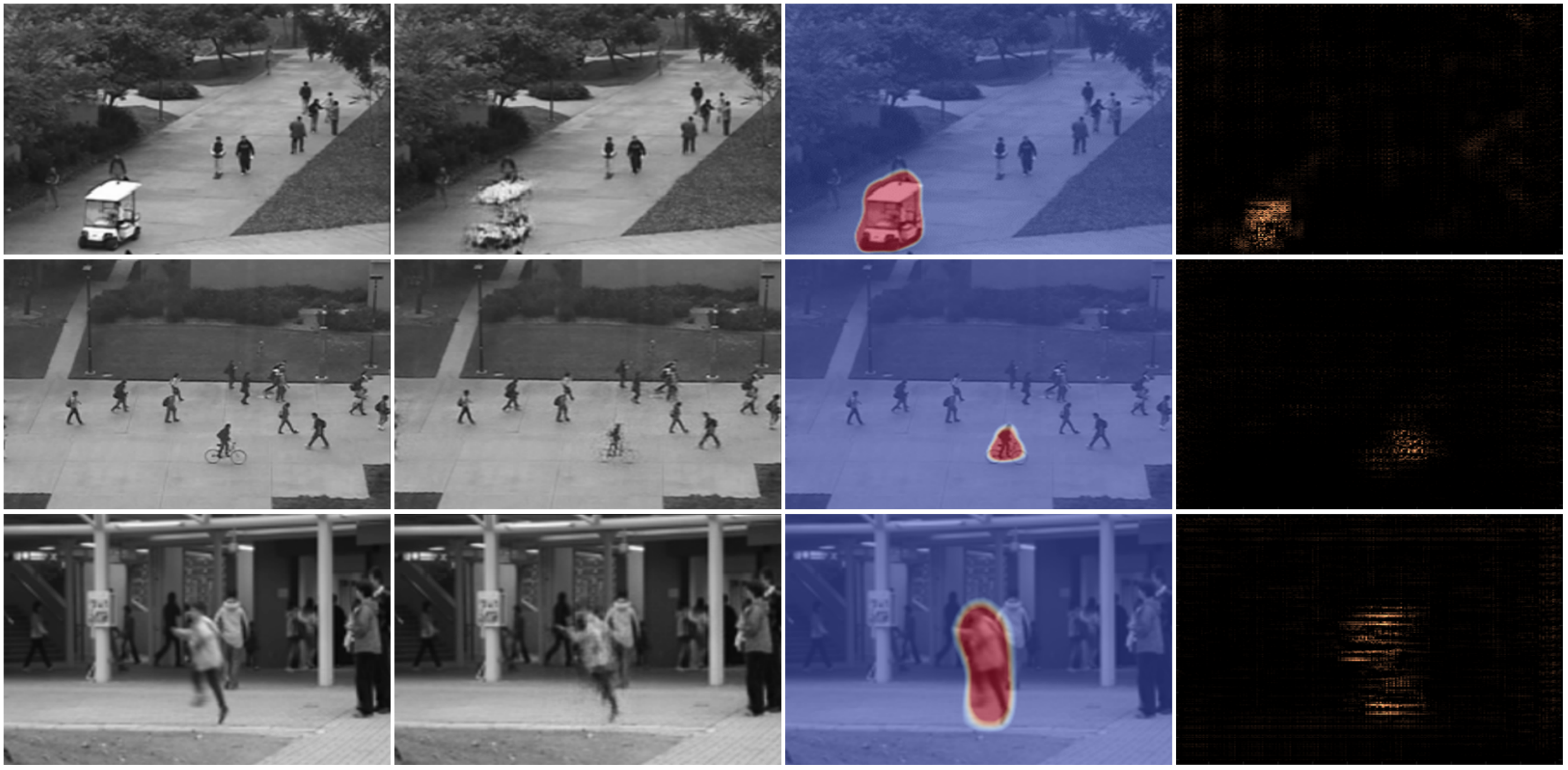}}
	\end{minipage}
\vspace{-0.5cm}
	\leftline{\qquad \ \ \  (a) \qquad \qquad \ \ \  (b) \qquad \qquad \ \ (c) \qquad \qquad \ \ \ (d)}\medskip
	\caption{Localization and visualization results of the abnormal events on UCSD Ped1 (first row), UCSD Ped2 (Second row), and Avenue (third row). (a) Real frame, (b) Generated frame, (c) Abnormality visualization by the generator, and (d) Abnormality visualization by the discriminator.}
	\label{fig:3}
	\vspace{-0.2cm}
	
\end{figure}

\subsection{Interpretation of the proposed networks}
\label{ssec:3.4}

Fig.~\ref{fig:3} shows localization and visualization of abnormal events detected by the proposed method for interpreting the networks. Fig.~\ref{fig:3}(c) shows visualization results obtained from the pixel-wise loss between the real frame and the generated frame. The proposed generator could detect the abnormal event regions such as a car, a bicycle, and a jumping person. Further, we visualized gradients of the discriminator using guided backpropagation \cite{22}. Fig.~\ref{fig:3}(d) shows gradients of activation at the output of the discriminator with respect to the input real sequences including the abnormal events. We observed that magnitude of the gradients was significantly high around the abnormal event regions. 

Through these results, we confirmed that each network performs abnormal event detection with the proper basis. The proposed networks could not only determine whether each frame is abnormal or not, but also detect where abnormal events occur at each frame.

\section{Conclusions}
\label{sec:4}

In this paper, we proposed the novel generative model based abnormal event detection method. To effectively represent the spatio-temporal features of normal patterns, the spatio-temporal generator and the spatio-temporal discriminator were trained in the adversarial way. As a result, the proposed method achieved competitive performance compared to the state-of-the-art methods. In particular, the proposed method far outperformed other methods on the dataset containing complex motion and frequent occlusions. Further, by visualizing the abnormal event regions detected by the proposed method, we could interpret how the STAN determine abnormal event at each scene.


\bibliographystyle{IEEEbib}
\bibliography{refs}

\begin{thebibliography}{10}

\bibitem{1}
Weiming Hu, Xuejuan Xiao, Zhouyu Fu, Dan Xie, Tieniu Tan, and Steve Maybank,
\newblock ``A system for learning statistical motion patterns,''
\newblock {\em IEEE transactions on pattern analysis and machine intelligence},
  vol. 28, no. 9, pp. 1450--1464, 2006.

\bibitem{3}
Shifu Zhou, Wei Shen, Dan Zeng, and Zhijiang Zhang,
\newblock ``Unusual event detection in crowded scenes by trajectory analysis,''
\newblock in {\em ICASSP}. IEEE, 2015, pp. 1300--1304.

\bibitem{4}
Vagia Kaltsa, Alexia Briassouli, Ioannis Kompatsiaris, Leontios~J
  Hadjileontiadis, and Michael~Gerasimos Strintzis,
\newblock ``Swarm intelligence for detecting interesting events in crowded
  environments,''
\newblock {\em IEEE transactions on image processing}, vol. 24, no. 7, pp.
  2153--2166, 2015.

\bibitem{5}
Andrei Zaharescu and Richard Wildes,
\newblock ``Anomalous behaviour detection using spatiotemporal oriented
  energies, subset inclusion histogram comparison and event-driven
  processing,''
\newblock in {\em ECCV}. Springer, 2010, pp. 563--576.

\bibitem{6}
Alex Krizhevsky, Ilya Sutskever, and Geoffrey~E Hinton,
\newblock ``Imagenet classification with deep convolutional neural networks,''
\newblock in {\em NIPS}, 2012, pp. 1097--1105.

\bibitem{7}
Dae~Hoe Kim, Wissam Baddar, Jinhyeok Jang, and Yong~Man Ro,
\newblock ``Multi-objective based spatio-temporal feature representation
  learning robust to expression intensity variations for facial expression
  recognition,''
\newblock {\em IEEE Transactions on Affective Computing}, 2017.

\bibitem{8}
Ian Goodfellow, Jean Pouget-Abadie, Mehdi Mirza, Bing Xu, David Warde-Farley,
  Sherjil Ozair, Aaron Courville, and Yoshua Bengio,
\newblock ``Generative adversarial nets,''
\newblock in {\em NIPS}, 2014, pp. 2672--2680.

\bibitem{9}
Mahmudul Hasan, Jonghyun Choi, Jan Neumann, Amit~K Roy-Chowdhury, and Larry~S
  Davis,
\newblock ``Learning temporal regularity in video sequences,''
\newblock in {\em CVPR}. IEEE, 2016, pp. 733--742.

\bibitem{10}
Jefferson~Ryan Medel and Andreas Savakis,
\newblock ``Anomaly detection in video using predictive convolutional long
  short-term memory networks,''
\newblock {\em arXiv preprint arXiv:1612.00390}, 2016.

\bibitem{23}
Xingjian Shi, Zhourong Chen, Hao Wang, Dit{-}Yan Yeung, Wai{-}Kin Wong, and
  Wang{-}chun Woo,
\newblock ``Convolutional lstm network: A machine learning approach for
  precipitation nowcasting,''
\newblock in {\em NIPS}, 2015, pp. 802--810.

\bibitem{11}
Mahdyar Ravanbakhsh, Moin Nabi, Enver Sangineto, Lucio Marcenaro, Carlo
  Regazzoni, and Nicu Sebe,
\newblock ``Abnormal event detection in videos using generative adversarial
  nets,''
\newblock in {\em ICIP}. IEEE, 2017.

\bibitem{12}
Du~Tran, Lubomir Bourdev, Rob Fergus, Lorenzo Torresani, and Manohar Paluri,
\newblock ``Learning spatiotemporal features with 3d convolutional networks,''
\newblock in {\em ICCV}. IEEE, 2015, pp. 4489--4497.

\bibitem{13}
Ashish Shrivastava, Tomas Pfister, Oncel Tuzel, Josh Susskind, Wenda Wang, and
  Russ Webb,
\newblock ``Learning from simulated and unsupervised images through adversarial
  training,''
\newblock in {\em CVPR}, 2017, pp. 2242--2251.

\bibitem{14}
Vijay Mahadevan, Weixin Li, Viral Bhalodia, and Nuno Vasconcelos,
\newblock ``Anomaly detection in crowded scenes,''
\newblock in {\em CVPR}. IEEE, 2010, pp. 1975--1981.

\bibitem{15}
Cewu Lu, Jianping Shi, and Jiaya Jia,
\newblock ``Abnormal event detection at 150 fps in matlab,''
\newblock in {\em ICCV}. IEEE, 2013, pp. 2720--2727.

\bibitem{24}
Diederik~P Kingma and Jimmy Ba,
\newblock ``Adam: A method for stochastic optimization,''
\newblock {\em arXiv preprint arXiv:1412.6980}, 2014.

\bibitem{16}
Jaechul Kim and Kristen Grauman,
\newblock ``Observe locally, infer globally: A space-time mrf for detecting
  abnormal activities with incremental updates,''
\newblock in {\em CVPR}. IEEE, 2009, pp. 2921--2928.

\bibitem{17}
Ramin Mehran, Alexis Oyama, and Mubarak Shah,
\newblock ``Abnormal crowd behavior detection using social force model,''
\newblock in {\em CVPR}. IEEE, 2009, pp. 935--942.

\bibitem{18}
Dan Xu, Yan Yan, Elisa Ricci, and Nicu Sebe,
\newblock ``Detecting anomalous events in videos by learning deep
  representations of appearance and motion,''
\newblock {\em Computer Vision and Image Understanding}, vol. 156, pp.
  117--127, 2017.

\bibitem{19}
Weixin Luo, Wen Liu, and Shenghua Gao,
\newblock ``Remembering history with convolutional lstm for anomaly
  detection,''
\newblock in {\em ICME}. IEEE, 2017, pp. 439--444.

\bibitem{20}
Radu Tudor~Ionescu, Sorina Smeureanu, Bogdan Alexe, and Marius Popescu,
\newblock ``Unmasking the abnormal events in video,''
\newblock in {\em ICCV}. IEEE, 2017, pp. 2914--2922.

\bibitem{21}
Weixin Luo, Wen Liu, and Shenghua Gao,
\newblock ``A revisit of sparse coding based anomaly detection in stacked rnn
  framework,''
\newblock in {\em ICCV}. IEEE, 2017, pp. 341 -- 349.

\bibitem{25}
Weixin Li, Vijay Mahadevan, and Nuno Vasconcelos,
\newblock ``Anomaly detection and localization in crowded scenes,''
\newblock {\em IEEE transactions on pattern analysis and machine intelligence},
  vol. 36, no. 1, pp. 18--32, 2014.

\bibitem{22}
Jost~Tobias Springenberg, Alexey Dosovitskiy, Thomas Brox, and Martin
  Riedmiller,
\newblock ``Striving for simplicity: The all convolutional net,''
\newblock {\em arXiv preprint arXiv:1412.6806}, 2014.

\end{thebibliography}
\label{sec:refs}

\end{document}